\documentclass[sigconf,nonacm]{acmart}
\usepackage{multirow}
\usepackage{makecell}
\usepackage{xspace}
\newcommand*{\eg}{e.g.\@\xspace}
\newcommand*{\ie}{i.e.\@\xspace}
\newcommand{\diag}{\mathop{\mathrm{diag}}}
\begin{document}

\title{Vision Transformer Pruning}

%\email{webmaster@marysville-ohio.com}
%\affiliation{%
%  \institution{Institute for Clarity in Documentation}
%  \streetaddress{P.O. Box 1212}
%  \country{USA}

%\author{%
%	Mingjian Zhu$^{1,2,3}$~~~~Kai Han$^{3}$~~~~Yehui Tang$^{3,4}$~~~~Yunhe Wang$^{3}$\\
%	$^1$Zhejiang University. $^2$Westlake University.\\
%	$^3$Noah's Ark Lab, Huawei Technologies. $^4$Peking University.\\
%	\texttt{zhumingjian@zju.edu.cn}, \texttt{\{kai.han,yunhe.wang\}@huawei.com} \\
%}

\author{Mingjian Zhu}
\affiliation{%
	\institution{Zhejiang University}
	\institution{Westlake University}  \country{}}
\email{zhumingjian@zju.edu.cn}

\author{Yehui Tang}
\affiliation{%
  \institution{Peking University}  \country{}}
\email{yhtang@pku.edu.cn}

\author{Kai Han}
\affiliation{%
	\institution{Noah's Ark Lab, Huawei Technologies}  \country{}}
\email{kai.han@huawei.com}

\begin{abstract}
Vision transformer has achieved competitive performance on a variety of computer vision applications. However, their storage, run-time memory, and computational demands are hindering the deployment to mobile devices. Here we present a vision transformer pruning approach, which identifies the impacts of dimensions in each layer of transformer and then executes pruning accordingly. By encouraging dimension-wise sparsity in the transformer, important dimensions automatically emerge. A great number of dimensions with small importance scores can be discarded to achieve a high pruning ratio without significantly compromising accuracy. The pipeline for vision transformer pruning is as follows: 1) training with sparsity regularization; 2) pruning dimensions of linear projections; 3) fine-tuning. The reduced parameters and FLOPs ratios of the proposed algorithm are well evaluated and analyzed on ImageNet dataset to demonstrate the effectiveness of our proposed method.
\end{abstract}

%\begin{CCSXML}
%<ccs2012>
%  <concept_id>10010520.10010553.10010562</concept_id>
%  <concept_desc>Computer systems organization~Embedded systems</concept_desc>
%  <concept_significance>500</concept_significance>
%  <concept_id>10010520.10010575.10010755</concept_id>
%  <concept_desc>Computer systems organization~Redundancy</concept_desc>
%  <concept_significance>300</concept_significance>
%  <concept_id>10010520.10010553.10010554</concept_id>
%  <concept_desc>Computer systems organization~Robotics</concept_desc>
%  <concept_significance>100</concept_significance>
%  <concept_id>10003033.10003083.10003095</concept_id>
%  <concept_desc>Networks~Network reliability</concept_desc>
%  <concept_significance>100</concept_significance>
%</ccs2012>
%\end{CCSXML}

%\ccsdesc[500]{Computer systems organization~Embedded systems}


\setcopyright{none}
\copyrightyear{}
\acmYear{}
\acmDOI{}
\acmConference{}
\acmBooktitle{}
\acmPrice{}
\acmISBN{}

\maketitle

\section{Introduction}
\label{sec:intro}
Recently, transformer~\cite{Att} has attracted much attention and shed light on various computer vision applications~\cite{han2020survey,liu2021swin,ipt,chen2020generative} such as image classification~\cite{vit,DeiT,han2021transformer}, object detection~\cite{detr,carion2020end,zhu2020deformable}, and image segmentation~\cite{wang2020end,wang2020max,hu2021istr}. However, most of the proposed transformer variants highly demand storage, run-time memory, and computational resource requirements, which impede their wide deployment on edge devices, e.g., robotics and mobile phones. Although massive effective techniques have been developed for compressing and accelerating convolutional neural networks (CNNs) including low-rank decomposition~\cite{tai2015convolutional,denton2014exploiting,lin2018holistic,lee2019learning,yu2017compressing}, quantization~\cite{gupta2015deep,rastegari2016xnor,yang2020searching,rastegari2016xnor,cai2017deep}, network pruning~\cite{tang2020scop,hassibi1993second,lecun1990optimal,wen2016learning,tang2020reborn}, and knowledge distillation~\cite{hinton2015distilling,tian2019contrastive,romero2014fitnets,luo2016face,li2020residual}, there still exists an urgency to develop and deploy efficient vision transformer.

Taking advantage of different designs~\cite{zafrir2019q8bert,tinybert,shen2020q,zhao2020investigation,zaheer2020big}, transformer can be compressed and accelerated to varying degrees. ALBERT~\cite{lan2019albert} reduces network parameter and speed up training time by decomposing embedding parameters into smaller matrices and enabling cross-layer parameter sharing. Star-Transformer~\cite{guo2019star} sparsifies the standard transformer by moving fully-connected structure to the star-shaped topology. Based on knowledge distillation techniques, the student networks in~\cite{sun2019patient,tinybert} learn from the logits in the larger pre-trained teacher networks. Some effective pruning algorithms have been proposed to reduce the attention heads~\cite{michel2019sixteen} or individual weights~\cite{gordon2020compressing}. The previous methods focus on compressing and accelerating the transformer for the natural language processing tasks. With the emergence of vision transformers such as ViT~\cite{vit}, PVT~\cite{wang2021pyramid}, and TNT~\cite{han2021transformer}, an efficient transformer is urgently need for computer vision applications. 

To address the aforementioned problems, we propose to prune the vision transformer according to the learnable importance scores. Inspired by the pruning scheme in network slimming~\cite{liu2017learning}, we add the learning importance scores before the components to be prune and sparsify them by training the network with $\mathcal{L}_1$ regulation. The dimensions with smaller importance scores will be pruned and the compact network can be obtained. Experimental results on the benchmark demonstrate the effectiveness of the proposed algorithm. Our vision transformer pruning (VTP) method largely compresses and accelerates the original ViT (DeiT) models. As the first pruning method for vision transformers, this work will provide a solid baseline and experience for future research.

\section{Approach}
\label{sec:method}

\subsection{Transformer}
The typical vision transformer architecture~\cite{Att,vit} consists of Multi-Head Self-Attention (MHSA), Multi-Layer Perceptron (MLP), layer normalization, activation function, and shortcut connection. MHSA is the characteristic component of transformer to perform information interaction among tokens. In particular, the input $X\in\mathbb{R}^{n\times d}$ is transformed to query $Q\in\mathbb{R}^{n\times d}$, key $K\in\mathbb{R}^{n\times d}$ and value $V\in\mathbb{R}^{n\times d}$ via fully-connected layers, where $n$ is the number of patches. $d$ is the embedding dimension. The self-attention mechanism is utilized to model the relationship between patches:
\begin{equation}
	\textit{Attention}(Q,K,V) = \textit{Softmax}\left({QK^\mathrm{T}}/{\sqrt{d}}\right)V.
\end{equation}
Finally, a linear transformation is applied to generate the output of MHSA:
\begin{equation}
	Y = X + \textit{FC}_{out}(\textit{Attention}(\textit{FC}_{q}(X),\textit{FC}_{k}(X),\textit{FC}_{v}(X))),
\end{equation}
where the layer normalization and activation function are omitted for simplification. As for the two-layer MLP, it can be formulated as%The number of parameters and FLOPs of MHSA are $4d^2$ and $4nd^2$+$2n^2d$, respectively. As for the two-layer MLP, it can be formulated as
\begin{equation}
	Z = Y + \textit{FC}_2(\textit{FC}_1(Y)).
\end{equation}
Intuitively, the widely-used fully-connected layers in the transformer lead to computation and storage burden.
%The hidden dimension is usually set as $4d$, so its parameters and FLOPs values are $8d^2$ and $8nd^2$, respectively. The parameters or FLOPs of layer normalization, activation function, and shortcut can be ignored compared to those of MHSA and MLP. A transformer block has about $12d^2$ parameters and $12nd^2$+$2n^2d$ FLOPs in total where MHSA and MLP occupy the vast majority.

\begin{figure*}[tp] 
	\centering
	\includegraphics[width=0.9\linewidth]{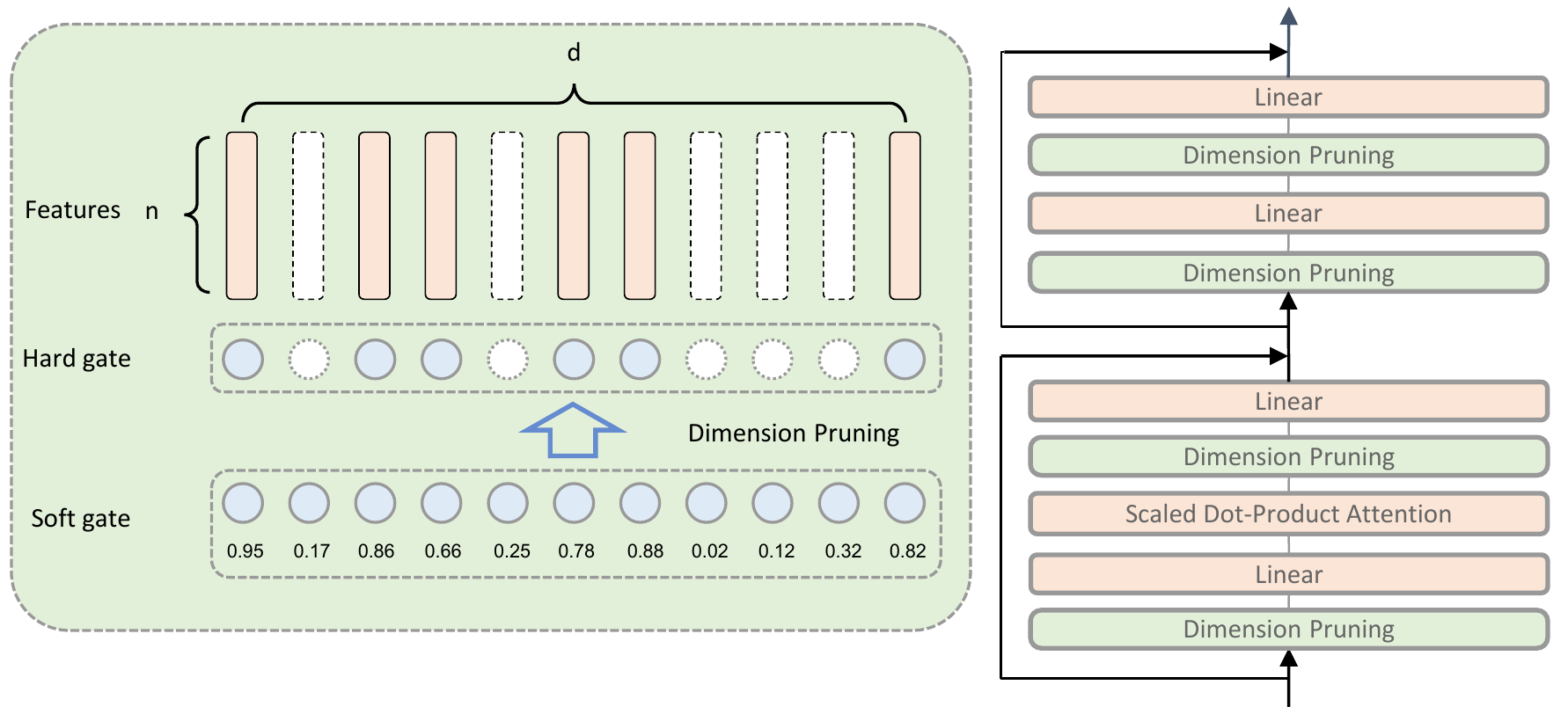}
	\vspace{-1em}
	\caption{Vision Transformer Pruning.}
	\label{fig:prune}
	\vspace{-1em}
\end{figure*}

\subsection{Vision Transformer Pruning}
To slim the transformer architecture, we focus on decreasing the FLOPs of MHSA and MLP. We propose to prune the dimension of the linear projection by learning their associated importance scores. For the features $X\in\mathbb{R}^{n\times d}$, where $n$ denotes the number of features that need to be pruned and $d$ denotes the dimension of each feature, we aim to preserve the generated important features and remove the useless ones from the corresponding components of linear projection. Suppose the optimal importance scores are $\mathbf{a}^*\in\{0,1\}^{d}$, that is, the scores for generated important features and their corresponding components are ones while the scores for useless ones are zeros. With the importance scores, we can obtain the pruned features:
\begin{equation}
	X^* = X\diag(\mathbf{a}^*).
\end{equation}
However, it's hard to optimize $\mathbf{a}^*$ in the neural network through a back-propagation algorithm due to its discrete values. Thus, we propose to relax $\mathbf{a}^*$ to real values as $\hat{\mathbf{a}}\in\mathbb{R}^{d}$. The soft pruned features are obtained as
\begin{equation}
	\hat{X} = X\diag(\hat{\mathbf{a}})
\end{equation}
Then, the relaxed importance scores $\hat{\mathbf{a}}$ can be learned together with the transformer network end-to-end. In order to enforce sparsity of importance scores, we apply $\ell_1$ regularization on the importance scores: $\lambda\|\hat{\mathbf{a}}\|_1$ and optimize it by adding on the training objective, where $\lambda$ is the sparsity hyper-parameter. After training with sparsity penalty, we obtain the transformer with some importance scores near zero. We rank all the values of regularized importance scores in the transformer and obtain a threshold $\tau$ according to a pre-defined pruning rate. With the threshold $\tau$, we obtain the discrete $\mathbf{a}^*$ by setting the values below the threshold as zero and higher values as ones:
\begin{equation}
	\mathbf{a}^* = \hat{\mathbf{a}} \geq \tau.
\end{equation}
After pruning according to the importance scores $\mathbf{a}^*$, the total pruned transformer is fine-tuned to diminish the accuracy drop. The above pruning procedure is denoted as
\begin{equation}
	X^* = \textit{Prune}(X).
\end{equation}

As shown in Figure~\ref{fig:prune}, we apply the pruning operation on all the MHSA and MLP blocks. The pruning process for them can be formulated as
\begin{align}
	Q, K, V &= \textit{FC}_{q}^{\prime}(\textit{Prune}(X)),\textit{FC}_{k}^{\prime}(\textit{Prune}(X)),\textit{FC}_{v}^{\prime}(\textit{Prune}(X)),\\
	Y &= X + \textit{FC}_{out}^{\prime}(\textit{Prune}(\textit{Attention}(Q, K, V))),\\
	Z &= Y + \textit{FC}_2^{\prime}(\textit{Prune}(\textit{FC}_1^{\prime}(\textit{Prune}(Y)))).
\end{align}
where $\textit{FC}_{q}^{\prime}, \textit{FC}_{k}^{\prime}, \textit{FC}_{v}^{\prime}, \textit{FC}_{out}^{\prime}, \textit{FC}_1^{\prime}$, and $\textit{FC}_2^{\prime}$ are pruned linear projection corresponding to the pruned features and $\mathbf{a}^*$. The proposed vision transformer pruning (VTP) method provides a simple yet effective way to slim vision transformer models. We hope that this work will serve as a solid baseline for future research and provide useful experience for the practical deployment of vision transformers.

\begin{table*}[htp]
	\small 
	\centering
	\caption{Ablation Study on ImageNet-100.}\label{tab:ablation_study}
	\vspace{-0.5em}
	\renewcommand{\arraystretch}{1.05}
	\setlength{\tabcolsep}{4pt}{
		%{
		\begin{tabular}{c|c|c|c|c|c|c}
			\toprule[1.0pt]
			Sparse Penalty              &Pruning Rate  & Params (M) & Params Reduced & FLOPs (B) &FLOPs Reduced & Top1 (\%) \\	
			\hline
			\multirow{4}{*}{0.0001}  &0.6   & 29.0 & $\downarrow$66.4\%  & 6.4  & $\downarrow$63.5\% & 90.00 \\
			&0.5   & 38.0 & $\downarrow$56.0\%  & 8.2  & $\downarrow$53.4\% & 91.46 \\
			&0.4   & 47.3 & $\downarrow$45.3\%  & 10.0 & $\downarrow$43.0\% & 92.58 \\
			&0.2   & 66.1 & $\downarrow$23.5\%  & 13.7 & $\downarrow$22.0\% & 93.54 \\
			\hline
			\multirow{4}{*}{0.00001} &0.6   & 28.2 & $\downarrow$67.4\%  & 6.3  & $\downarrow$64.4\% & 89.88 \\
			&0.5   & 37.5 & $\downarrow$56.6\%  & 8.1  & $\downarrow$54.0\% & 91.40 \\
			&0.4   & 47.1 & $\downarrow$45.5\%  & 10.0 & $\downarrow$43.2\% & 92.38 \\
			&0.2   & 66.1 & $\downarrow$23.5\%  & 13.7 & $\downarrow$22.0\% & 93.44 \\
			\bottomrule[1pt]
		\end{tabular}
	}
	\vspace{-0.5em}
\end{table*}

\begin{table*}[htp]
	\small 
	\centering
	\caption{Results on ImageNet-100.}\label{tab:ImageNet-100}
	\vspace{-0.5em}
	\renewcommand{\arraystretch}{1.05}
	\setlength{\tabcolsep}{8pt}{
		\begin{tabular}{l|c|c|c|c}
			\toprule[1.5pt]
			Model       & Params (M) & FLOPs (B) & Top1 (\%) & Top5 (\%) \\
			%\hline
			%Deit-S (Baseline)     & running & running & running & running  \\
			%VTP-S-Sparse(0.0001)  & running & running & running & running \\
			%VTP-S (Pruned-A)      & running &running & running & running \\
			%VTP-S (Pruned-B)      & running & running & running & running \\
			\hline
			Deit-B (Baseline)  & 86.4 & 17.6 & 94.50 & 98.94 \\
			%VTP-B-Sparse(0.0001)  & 86.4 & 17.6 & 93.50 & 98.34 \\
			VTP (20\% pruned)  & 66.1 & 13.7 & 93.54 & 98.36 \\
			VTP (40\% pruned)  & 47.3 & 10.0 & 92.58 & 98.04 \\
			\bottomrule[1pt]
		\end{tabular}
	}
	\vspace{-0.5em}
\end{table*}

\begin{table*}[htp]
	\small 
	\centering
	\caption{Results on ImageNet-1K.}\label{tab:ImageNet-1K}
	\vspace{-0.5em}
	\renewcommand{\arraystretch}{1.05}
	\setlength{\tabcolsep}{8pt}{
		\begin{tabular}{l|c|c|c|c}
			\toprule[1.5pt]
			Model       & Params (M) & FLOPs (B) & Top1 (\%) & Top5 (\%) \\
			\midrule
			\emph{CNN based} & & & & \\
			%ResNet-50 & 25.6 & 4.1 & 76.2 & 92.9 \\
			ResNet-152 & 60.2 & 11.5 & 78.3 & 94.1 \\
			%RegNetY-8GF & 39.2 & 8.0 & 79.9 & - \\
			RegNetY-16GF & 83.6 & 15.9 & 80.4 & - \\
			%EfficientNet-B3 & 12.0 & 1.8 & 81.6 & 94.9 \\
			%EfficientNet-B4 & 19.0 & 4.2 & 82.9 & 96.4 \\
			\midrule
			\emph{Transformer based} & & & & \\
			ViT-B/16  & 86.4 & 55.5 &  77.9 & - \\
			%ViT-L/16  & 304.3 & 63.6 &  76.5 & - \\
			%DeiT-S  & 22.1 & 4.6 & 79.8 & - \\
			DeiT-B (Baseline)   & 86.4 & 17.6 & 81.8 & - \\
			VTP (20\% pruned)  & 67.3 & 13.8 & 81.3 & 95.3 \\
			VTP (40\% pruned)   & 48.0 & 10.0 & 80.7 & 95.0 \\
			
			\bottomrule[1pt]
		\end{tabular}
	}
	\vspace{-0.5em}
\end{table*}

\section{Experiments}
In this section, we verify the effectiveness of the proposed VTP methods to prune vision transformer models on ImageNet dataset.

\subsection{Datasets}

\paragraph{ImageNet-1K.}
ImageNet ILSVRC2012 dataset~\cite{russakovsky2015imagenet} is a large-scale image classification dataset including 1.2 million images for training and 50,000 validation images belonging to 1,000 classes. The common data augmentation strategy in DeiT~\cite{DeiT} is adopted for model development, including Rand-Augment~\cite{cubuk2020randaugment}, Mixup~\cite{zhang2017mixup}, and CutMix~\cite{yun2019cutmix}.

\paragraph{ImageNet-100.}
ImageNet-100 is collected as a subset of ImageNet-1K. We first randomly sampled 100 classes and their corresponding images for training and validation. We adopt the same data augmentation strategy for ImageNet-100 as ImageNet-1K.

\subsection{Implementation Details}

\paragraph{Baseline.} We evaluate our pruning method on a popular vision transformer implementation, \ie, DeiT-base~\cite{DeiT}. In our experiments, a 12-layer transformer with 12 heads and 768 embedding dimensions is evaluated on both ImageNet-1K and Imagenet-100. For a fair comparison, we utilize the official implementation of DeiT and do not use techniques like distillation. On the ImageNet-1K, we take the released model of DeiT-base as the baseline. We finetune the model on the ImageNet-1K using batch size 64 for 30 epochs. The initial learning rate is set to ${6.25 \times 10^{-7}}$. Following Deit~\cite{DeiT}, we use AdamW~\cite{loshchilov2018fixing} with cosine learning rate decay strategy to train and finetune the models.

\paragraph{Training with Sparsity Regularization and Pruning.} Based on the baseline model, we train the vision transformer with $\ell_1$ regularization using different sparse regularization rates. We select the optimal sparse regularization rate (\ie 0.0001) on Imagenet-100 and apply it on ImageNet-1K. The learning rate for training with sparsity is ${6.25 \times 10^{-6}}$ and the number of epochs is 100. The other training setting follows the baseline model.  After sparsity, we prune the transformer by setting different pruning thresholds and the threshold is computed by the predefined pruning rate, \eg, 0.2.

\paragraph{Finetuning.} We finetune the pruned transformer with the same optimization setting as in training, except for removing the $\ell_1$ regularization.

\subsection{Results and Analysis}
\paragraph{Imagenet-100 Experiments and Ablation Study.} 
We firstly conduct ablation studies on Imagenet-100, as shown in Table~\ref{tab:ablation_study}. From the results, the amount of pruning rate matches the ratio of parameters saving and FLOPs saving. For example, when we prune 40\% dimensions of the models trained with 0.0001 sparse rate, the parameter saving is 45.3\% and the FLOPs saving is 43.0\%. We can see that the Parameters and FLOPs drop while the accuracy maintains. Besides, the sparse ratio does not highly influence the effectiveness of the pruning method. In Table~\ref{tab:ImageNet-100}, we compare the baseline model with two VTP models, \ie, 20\% pruned and 40\% pruned models. The accuracy drops slightly with large FLOPs decrease. When we prune 20\% dimensions, 22.0\% FLOPs are saved and the accuracy drops by 0.96\%. When we prune 40\% dimensions, 45.3\% FLOPs are saved and the accuracy drops by 1.92\%.

\vspace{-0.5em}
\paragraph{Imagenet-1K Experiments.}
We also evaluate the proposed VTP method on the large-scale ImageNet-1K benchmark. The results are shown in Table~\ref{tab:ImageNet-1K}. Compared to the base model DeiT-B, the accuracy of VTP only decreases by 1.1\% when 40\% dimensions are pruned. The accuracy only drops by 0.5\% while 20\% dimensions are pruned. The effectiveness of VTP can be generalized to large-scale datasets.

\section{Conclusion}
In this paper, we introduce a simple yet efficient vision transformer pruning method. $\mathcal{L}_1$ regulation is applied to sparse the dimensions of the transformer and the important dimensions appear automatically. The experiments conducted on Imagenet-100 and ImageNet-1K demonstrate that the pruning method can largely reduce the computation costs and model parameters while maintaining the high accuracy of original vision transformers. In the future, the important components such as the number of heads and the number of layers can also be reduced with this method, which is a promising attempt to further compress vision transformers.


\bibliographystyle{ACM-Reference-Format}
\bibliography{sample-base}

%\appendix

\end{document}